\documentclass[francais]{gretsi}
 \usepackage[numbers]{natbib}
\usepackage[utf8]{inputenc}

\usepackage[T1]{fontenc}

\usepackage{amsmath, amssymb, amsfonts}

 \usepackage{subcaption}
 \usepackage{multirow}
 \usepackage{xcolor}

 \usepackage{url} 
 \usepackage[mode=buildnew]{standalone}

\newcommand{\meteorix}[0]{{\sc Meteorix}\xspace}
\newcommand{\source}[0]{{\sc Source}\xspace}
\newcommand{\fripon}[0]{{\sc Fripon}\xspace}
\newcommand{\rpi}[0]{Raspberry Pi 4\xspace}

\newcommand{\new}[1]{\color{black}#1\color{black}}

\begin{document}

\titre{Une nouvelle application de détection de météores\\ robuste aux mouvements de caméra}

   \auteurs{\auteur{C.}{Ciocan}{clara.ciocan@lip6.fr}{1}
           \auteur{M.}{Kandeepan}{mathuran.kandeepan@lip6.fr}{1}
           \auteur{A.}{Cassagne}{adrien.cassagne@lip6.fr}{1}
           \auteur{J.}{Vaubaillon}{jeremie.vaubaillon@imcce.fr}{2}
           \auteur{F.}{Zander}{Fabian.Zander@usq.edu.au}{3} 
           \auteur{L.}{Lacassagne}{lionel.lacassagne@lip6.fr}{1}}
   \affils{\affil{1}{LIP6, Sorbonne Université, CNRS, France}
     \affil{2}{IMCCE, CNRS, Observatoire de Paris, PSL Université, Sorbonne Université, Université de Lille 1, France}
     \affil{3}{University of Southern Queensland, Australia}}

\resume{ Cet article présente un nouvel outil pour la détection automatique des météores nommé \textit{Fast Meteor Detection Toolbox} (FMDT) qui est capable de détecter des météores entrant dans l'atmosphère terrestre en analysant des vidéos acquises avec des caméras embarquées à bord d'un ballon-sonde ou stabilisées dans un avion. Le défi est de concevoir une chaîne de traitement composée d'algorithmes simples tout en étant robustes à une grande variabilité de vidéos et en respectant des contraintes de consommation (10 W) et de temps réel (25 images par seconde).

}
\abstract{
This article presents a new tool for the automatic detection of meteors. \textit{Fast Meteor Detection Toolbox} (FMDT) is able to detect meteor sightings by analyzing videos acquired by cameras onboard weather balloons or within airplane with stabilization. The challenge consists in designing a processing chain composed of simple algorithms, that are robust to the high fluctuation of the videos and that satisfy the constraints on power consumption (10 W) and real-time processing (25 frames per second).
}

\maketitle
\begin{sloppypar}
\section{Introduction}
\vspace{-0.1cm}

    Un météore est un phénomène lumineux produit par l'entrée d'un corps céleste dans l'atmosphère terrestre.
    Ce phénomène est d'un intérêt majeur pour les astronomes : son observation et sa variation de luminosité permet d'en déduire sa composition et  son origine afin d'en apprendre plus sur la formation du Système Solaire. La collecte de ces séquences vidéos génère de gros volumes d'images à analyser.

    Dans cet article, une nouvelle application, nommée \emph{Fast Meteor Detection Toolbox} (FMDT)\footnote{https://github.com/alsoc/fmdt/}, est proposée pour du traitement temps réel embarqué à bord  de ballons-sondes~\cite{ocana2019balloon} ou d'avions~\cite{Vaubaillon2023_short}. La caméra est donc en mouvement, ce qui complexifie le traitement par rapport à une observation depuis le sol. La chaîne de traitement FMDT doit donc réaliser un compromis entre complexité algorithmique, temps de traitement, consommation et qualité de la détection. 

    

    La section 2 présente l'état de l'art des applications de détection de météores. La section 3 décrit la chaîne de traitement et le banc de validation. L'analyse des résultats obtenus est détaillée dans la section 4.

\vspace{-0.2cm}
\section{État de l'art}
\vspace{-0.1cm}

    Depuis les années 1990, plusieurs chaînes de traitement d'images ont été développées. Le traitement est effectué en trois étapes : pré-traitement des images, détection de régions d'intérêt et leur classification (météores, villes, orages, ...). 
    
    Les applications {\sc MetRec}~\cite{metrec} et {\sc MeteorScan}~\cite{grl} permettent une détection en temps réel avec une probabilité de détection supérieure à 80\%~\cite{MolGur}. Leur pré-traitement inclut une soustraction des pixels stationnaires entre deux images. Puis une transformée de Hough~\cite{duda1972use} est réalisée pour détecter les potentiels météores. Un pré-traitement similaire est utilisé dans l'application  {\sc FreeTure}~\cite{2021ascl.soft04011A} du projet \fripon~\cite{Colas2020_FRIPON_short} qui utilise plus de 250 caméras orientées vers le ciel, réparties dans plusieurs pays. 

    Récemment, des réseaux de neurones profonds~\cite{galindo,  dlgural, cecil} ont permis d'atteindre des taux de détection supérieurs à 95\%. Ces résultats sont à relativiser car  d'une part, la caméra est fixe -- ce qui est un cas plus simple qu'une caméra en mouvement -- et que d'autre part, les CPU (Intel i7-6850K) et GPU (Nvidia Quadro 4000) utilisés ont tous deux un TDP de 140 W. Enfin, un seul de ces projets mentionne avoir atteint une cadence proche du temps réel (21 images/s)~\cite{cecil}. 
    A contrario, FMDT est conçu pour des SoC consommant moins de 10 Watt.



    Les applications déployées au sol sont aussi dépendantes des conditions météorologiques (nuages, éclairs) et de la pollution lumineuse.  La première mission depuis l'espace, pour pallier ce problème, est le projet {\sc Meteor} à bord de l'ISS \cite{Arai2014}.
    \meteorix~\cite{Rambaux2019_ESA_short}\cite{Millet2022_Meteorix_COSPAR} et \source~\cite{liegibel2022meteor} sont des projets de nanosatellite universitaire actuellement en phase de définition où la caméra pointe vers la Terre. Cela nécessite un algorithme de flot optique pour calculer la vitesse apparente en tout point \cite{Petreto2018_SIMD_GPU_OF_DASIP} et faire la différence entre la vitesse de défilement des tâches lumineuses à la surface de la Terre des autres zone lumineuses comme les nuages éclairés par la Lune, les éclairs et les météores.
    Dans le cas d'un ballon-sonde ou d'un avion, la caméra pointe au limbe (axe tangentiel à la Terre). Les scènes sont majoritairement sur un fond noir, avec des étoiles et une partie de l'atmosphère où sont visibles les météores.  Il n'est pas nécessaire de calculer le flot.

\vspace{-0.2cm}
\section{Chaîne de traitement FMDT}
\vspace{-0.1cm}

\begin{figure*}[htp]
  \begin{center}
  \begin{subfigure}{\textwidth}
  \centering
  \includegraphics[scale=0.45,keepaspectratio]{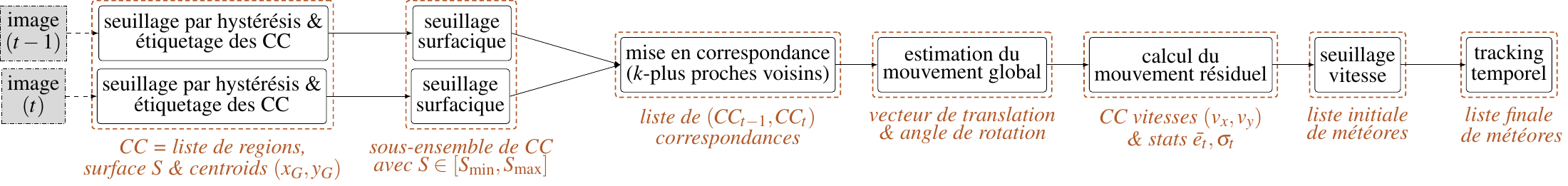}
  \subcaption{Chaîne de traitement principale basée sur le calcul du mouvement entre deux images $I_{t - 1}$ et $I_t$ et  \it{tracking} temporel.}
  \label{fig:chaine1}
  \vspace{0.2cm}
  \end{subfigure}
  \begin{subfigure}{\textwidth}
  \centering
  \includegraphics[scale=0.45,keepaspectratio]{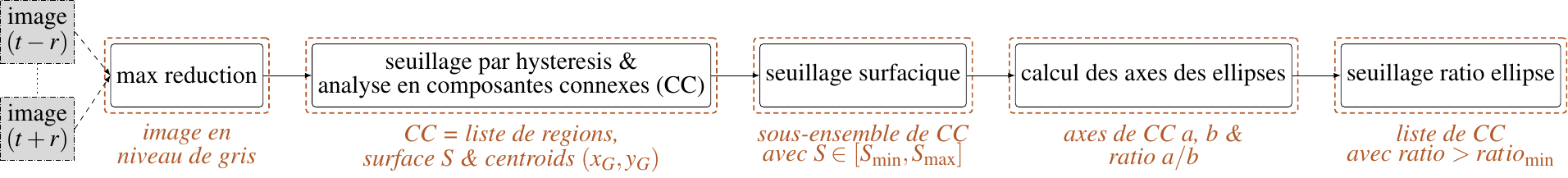}
  \subcaption{Chaîne de traitement secondaire basée sur la \emph{max-réduction} d'images entre $I_{t - r}$, ..., $I_t$, ..., et $I_{t + r}$ et calcul d'ellipse.}
  \label{fig:chaine2}
  \end{subfigure}
  \vspace{-0.5cm}
  \caption{Synoptique de la chaîne de traitement principale de FMDT et d'une extension pour la détection d'ellipses.}
  \label{fig:chaines}
  \end{center}
\end{figure*}



La chaîne de traitement proposée repose sur l'hypothèse que les météores  sont moins nombreux que les étoiles (ou les planètes brillantes). Ainsi, extraire les tâches lumineuses et faire une mise en correspondance doit permettre de recaler les objets ``à l'infini''. Ceux mal recalés et ayant une trajectoire linéaire sont les météores.

\vspace{-0.1cm}
\subsection{Aspect spatial}
\vspace{-0.1cm}


\new{Le seuillage par hystérésis de la chaine principale (Fig ~\ref{fig:chaine1}) est réalisé par l'intersection ensembliste de deux ensembles de composantes connexes (CC) produits par l'étiquetage et l'analyse en composantes connexes (ECC \& ACC) des images binarisées aux seuils haut et bas de l'hystéresis ($\tau_\text{high}$ et $\tau_\text{low}$). ECC et ACC sont basés sur le Light Speed Labeling (LSL) \cite{Lacassagne2009_LSL_ICIP}, qui est un étiquetage combinant traitement par segment et compression RLE, permettant d'être très rapide en ACC \cite{Cabaret2014_CCL_SIPS}.}

    \begin{figure}[b]
\begin{subfigure}{\linewidth}
    \centering
    \begin{subfigure}{0.15\linewidth}
        \includegraphics[width=\linewidth]{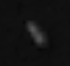}
        \caption{}
    \end{subfigure}
    \quad
    \begin{subfigure}{0.15\linewidth}
        \includegraphics[width=\linewidth]{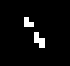}
        \caption{}
    \end{subfigure}
    \quad
    \begin{subfigure}{0.15\linewidth}
        \includegraphics[width=\linewidth]{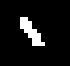}
        \caption{}
   
    \end{subfigure}

\end{subfigure}\hfill
\vspace{-0.4cm}
 \caption{Météore et sur-segmentation : (a) météore en niveau de gris, (b) seuillage simple à 75, (c) seuillage par hystérésis $\tau_\text{low}=55$, $\tau_\text{high}=70$.}
 \label{fig:sursegmentation}
\end{figure}

\begin{figure}[t]
\begin{subfigure}{\linewidth}
    \centering
    \begin{subfigure}{\linewidth}
        \includegraphics[width=\linewidth]{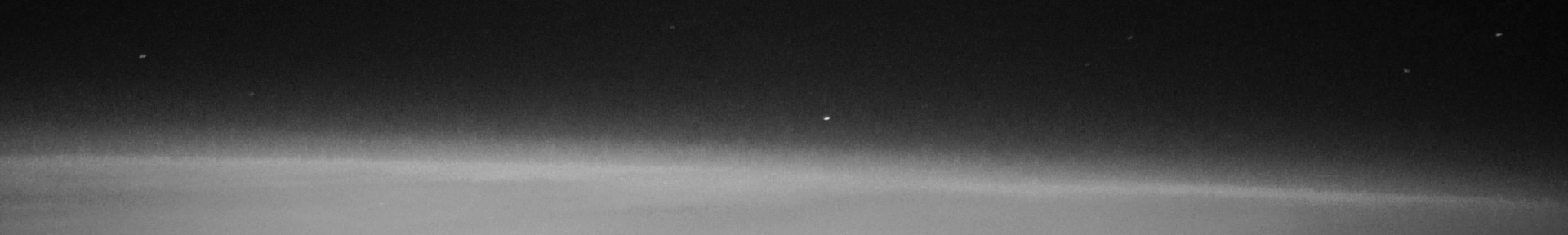}
        \subcaption{Atmosphère terrestre en niveau de gris.}
        \vspace{0.2cm}
    \end{subfigure}
    \begin{subfigure}{\linewidth}
        \includegraphics[width=\linewidth]{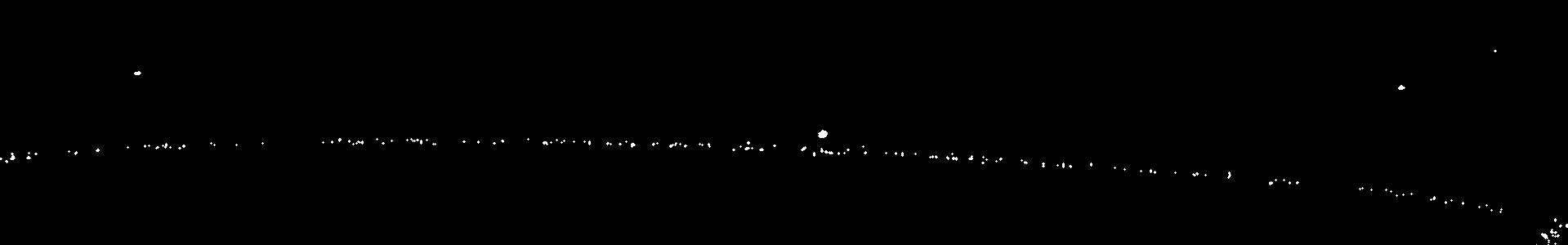}
        \subcaption{Seuillage simple à 75: des pixels de l'atmosphère sont détectés et génèrent 184 régions.}
        \vspace{0.2cm}
    \end{subfigure}
    \begin{subfigure}{\linewidth}
        \includegraphics[width=\linewidth]{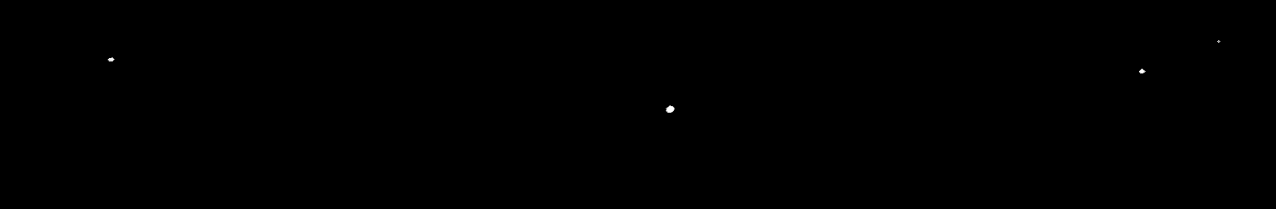}
        \subcaption{Seuillage par hystérésis $\tau_\text{low}$=55, $\tau_\text{high}$=70: 4 étoiles détectées.}
    \end{subfigure}

\end{subfigure}\hfill
\vspace{-0.2cm}
\caption{Comparaison du seuillage simple et du seuillage par hystérésis (extrait des Géminides).}
\label{fig:atm}
\vspace{-0.4cm}
\end{figure}

Concernant la sur-segmentation des météores, le seuillage par hystérésis est plus robuste qu'un seuillage simple (Fig.~\ref{fig:sursegmentation}) et, combiné à un filtrage sur la taille des régions $\left[ S_\text{min}, S_\text{max}\right]$, il permet de limiter les faux positifs dûs à l'atmosphère (Fig.~\ref{fig:atm}). 

Une première mise en correspondance par $k$-plus-proches-voisins \cite{Mucherino2009} permet une première estimation du mouvement global (translation, puis rotation) entre deux images successives à l'instant $t$. 
\new{Cet algorithme associe à la CC de l'instant $t$ la CC à $t+1$ la plus proche n'ayant pas été associée parmi les $k$ ($k=3$ par défaut) CC les plus proches. De plus, les associations sont faites avec une contrainte sur le ratio des surfaces, ce qui permet de faire la bonne association lorsque deux CC (par exemple une étoile et un météore) se trouvent à proximité. } L'erreur moyenne de recalage $\bar{e}_t$ et son écart type $\sigma_t$ servent à détecter les objets dont l'erreur de recalage $e_k$ est plus grande que la moyenne. Un seuil à 1 écart-type donne de bons résultats  ($|e_k-\bar{e_t}| > \sigma_t$). Un second recalage est alors effectué en ne prenant que les régions correspondant à des étoiles. À titre d'exemple, la moyenne géométrique du premier recalage du cluster de météores dans~\cite{Vaubaillon2023_short}, est de 0.91 pixel, tandis que celle du second recalage est de 0.18 pixel.

\vspace{-0.1cm}
\subsection{Aspect temporel : \textit{tracking}}
\vspace{-0.1cm}

Un tracking par morceau (avec une contrainte sur l'angle) permet de déterminer les trajectoires des régions. Il réduit le nombre de faux positifs, car les météores sont des phénomènes de courte durée, généralement inférieure à quelques secondes, avec une trajectoire \emph{rectiligne}, tandis que les régions correspondant à des étoiles persistent sur la totalité de la séquence. \new{Une régression linéaire des moindres carrés est faite pour chaque piste et confirme l'hypothèse précédente. La distance moyenne des points à la droite est de 0.19 pixel}. Enfin, l'algorithme de \textit{tracking} permet d'extrapoler les pistes et de réacquérir une région si le temps d'extrapolation maximum de 3 images n'est pas dépassé.

\begin{figure*}[t]
   \centering
   \includegraphics[width=1\columnwidth,keepaspectratio]{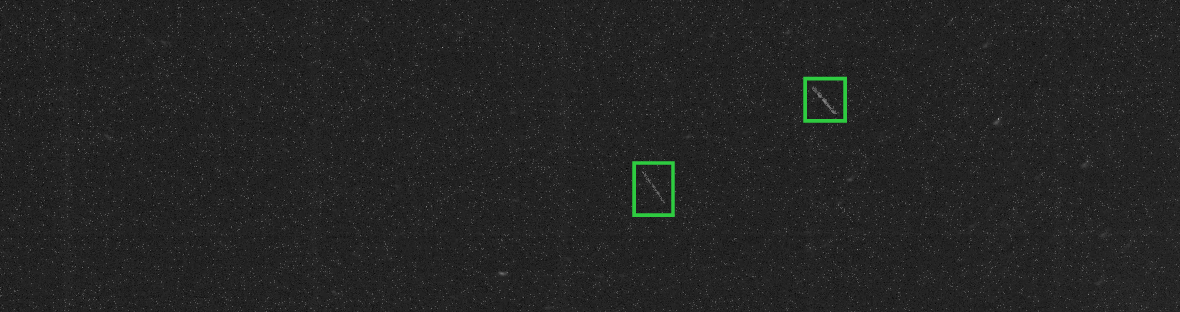}
   \quad
   \includegraphics[width=1\columnwidth,keepaspectratio]{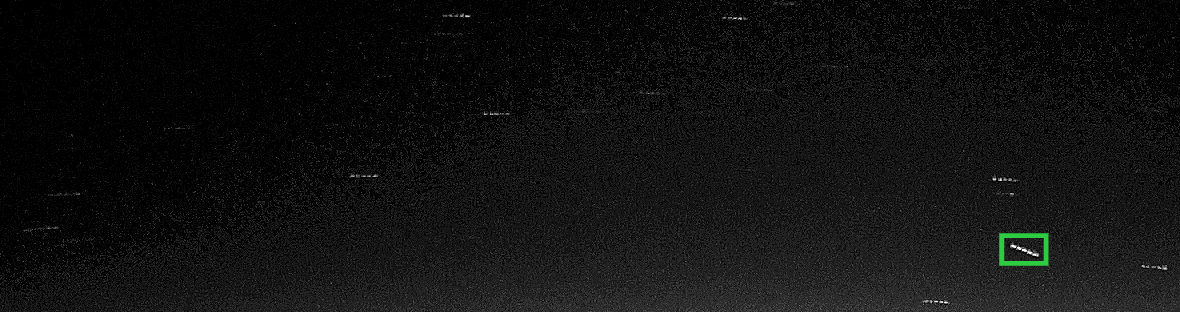}
   \caption{Max-réduction sur 5 images : les météores des $\tau$-Herculides génèrent des ellipses plus longues que l'étalement des étoiles (à gauche) tandis que pour les Géminides (à droite) les étoiles génèrent des traînées comparables à celles des météores. Les météores sont encadrés en vert.}
   \label{fig:maxred}
\end{figure*}

\vspace{-0.1cm}
\subsection{Classification par détection d'ellipse}
\vspace{-0.1cm}

Si l'on fait l'hypothèse que le mouvement apparent des météores est supérieur à celui des étoiles (dû au mouvement du porteur), alors, rechercher des CC en forme d'ellipse peut améliorer la détection. Afin de contenir les problèmes de sur-segmentation une \emph{max-réduction} -- temporelle de rayon $r$ en niveau de gris -- des images d'indice [$t-r$, $t+r$] est réalisée pour créer une image composite en amont du seuillage par hystérésis (Fig.~\ref{fig:chaine2}). Cela se fait efficacement, car l'encodage RLE du LSL fait que le calcul des moments statistiques d'ordre 2  est proportionnel au nombre de segments et non au nombre de pixels qui compose la CC.Ainsi, il est possible de calculer lu grand et lu petit rayon $(a,b)$ de l'ellipse, ainsi que son ratio d'aplatissement $\rho=a/b$.

\vspace{-0.3cm}
\section{Expérimentations}
\vspace{-0.1cm}
\subsection{Banc de test}
\vspace{-0.2  cm}

\begin{figure}[b]
\begin{subfigure}{\linewidth}  
    \centering
    \begin{subfigure}{0.4\linewidth}
        \includegraphics[width=\linewidth]{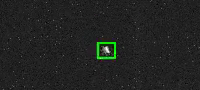}
        \caption{météore (vrai positif)}
    \end{subfigure}
    \quad
    \begin{subfigure}{0.4\linewidth}
        \includegraphics[width=\linewidth]{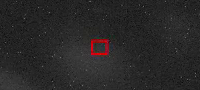}
        \caption{bruit (faux positif)}
    \end{subfigure}

\end{subfigure}\hfill
\vspace{-0.4cm}
 \caption{Classification des CC.}
 \label{fig:exemple}
\end{figure}

Un banc de test a été conçu pour qualifier la chaîne de traitement. Il est composé de vidéos de deux campagnes d'observation.
D'une part de vidéos des Géminides 2016~\cite{ocana2019balloon} qui sont acquises depuis un ballon sonde non stabilisé et qui subit les courants de haute altitude qui font osciller la nacelle. Le tangage et le roulis se traduisent par une translation verticale et une rotation, mais le lacet -- plus problématique -- se traduit par une importante translation horizontale. D'autre part, des vidéos des $\tau$-Herculides 2022, qui sont acquises depuis un avion; certaines des caméras sont stabilisées avec une \textit{gimbal} 3 axes (tangage, lacet, roulis), tandis que d'autres sont juste fixées à l'avion. 

\vspace{-0.1cm}
\subsection{Analyse des résultats}
\vspace{-0.1cm}


    La Table~\ref{stats} présente les résultats de détection sur le banc de test.
    Sur les 257 météores présents, FMDT en détecte 217. Le taux de détection, par rapport aux météores, est de 84\%. Les météores sont présents sur 3078 images et FMDT les détecte sur 1932 images. Cela représente un taux de détection, par rapport aux images, de 64\%.
     \new{Les paramètres des algorithmes ont été choisis en se basant sur les vérités terrains pour maximiser le nombre de détections correctes (objectif scientifique) quitte à augmenter les faux positifs (FP) (Fig. ~\ref{fig:exemple})}. Il est du même ordre que les vrais positifs, ce qui est acceptable. De plus, la chaîne de traitement s'exécute en 36 ms (26 images/s) sur un \rpi, tout en ne consommant que 6 Watts. Les contraintes d'embarquabilité sont donc satisfaites.

    À noter que les circonstances distinctes et le mouvement fluctuant présent sur les vidéos capturées ont permis de tester et de confirmer l'adaptabilité de la chaîne de traitement, due au nombre important de paramètres modifiables de l'outil. 

\begin{figure}[t]
\vspace{-0.5cm}
\begin{subfigure}{\linewidth}
    \centering
    \begin{subfigure}{\linewidth}
        \centering
        \includegraphics[width=0.95\columnwidth,keepaspectratio]{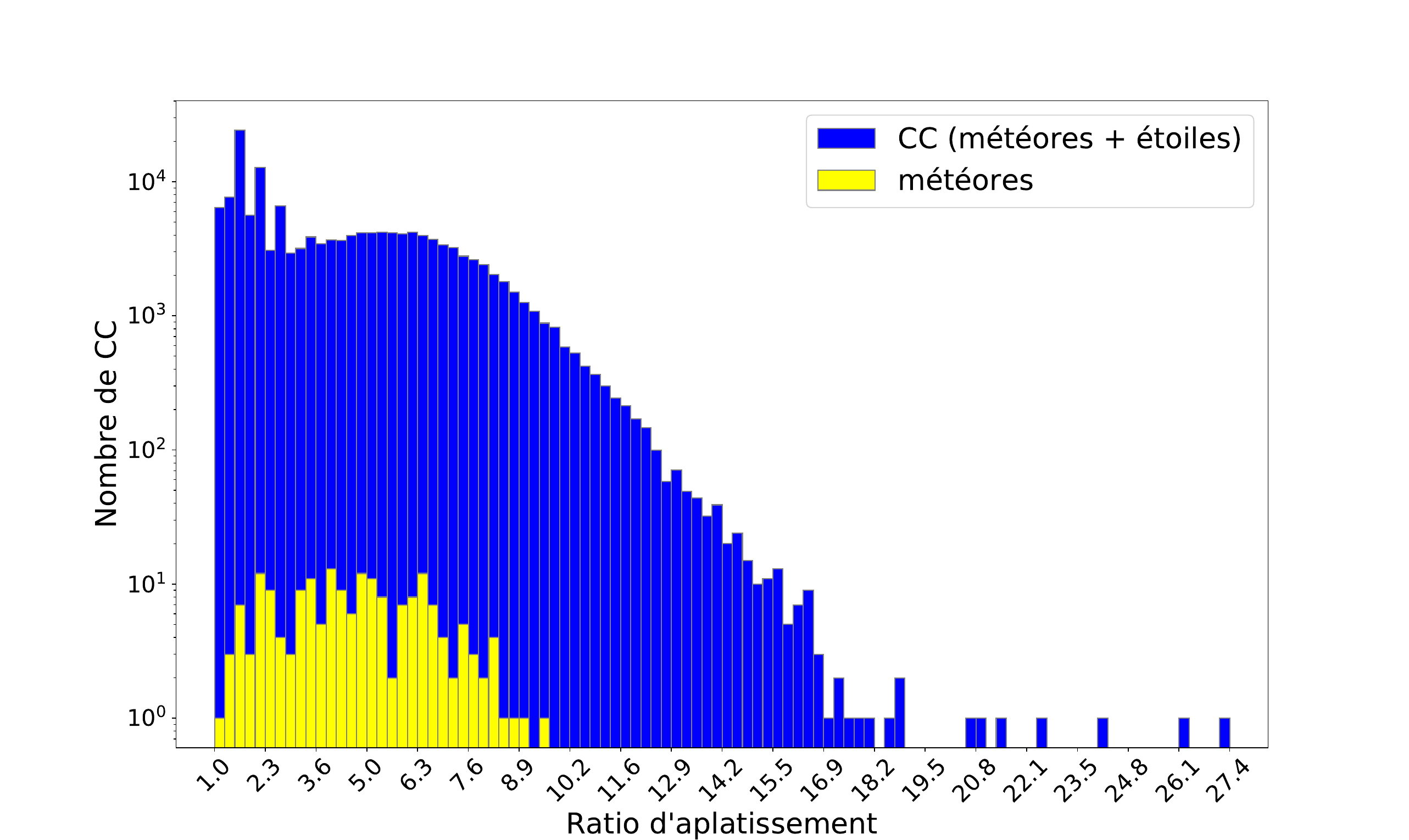}
        \vspace{-0.2cm}
        \subcaption{Exemple de séquence des Géminides.}
    \end{subfigure}
    \begin{subfigure}{\linewidth}
        \centering
        \includegraphics[width=0.95\columnwidth,keepaspectratio]{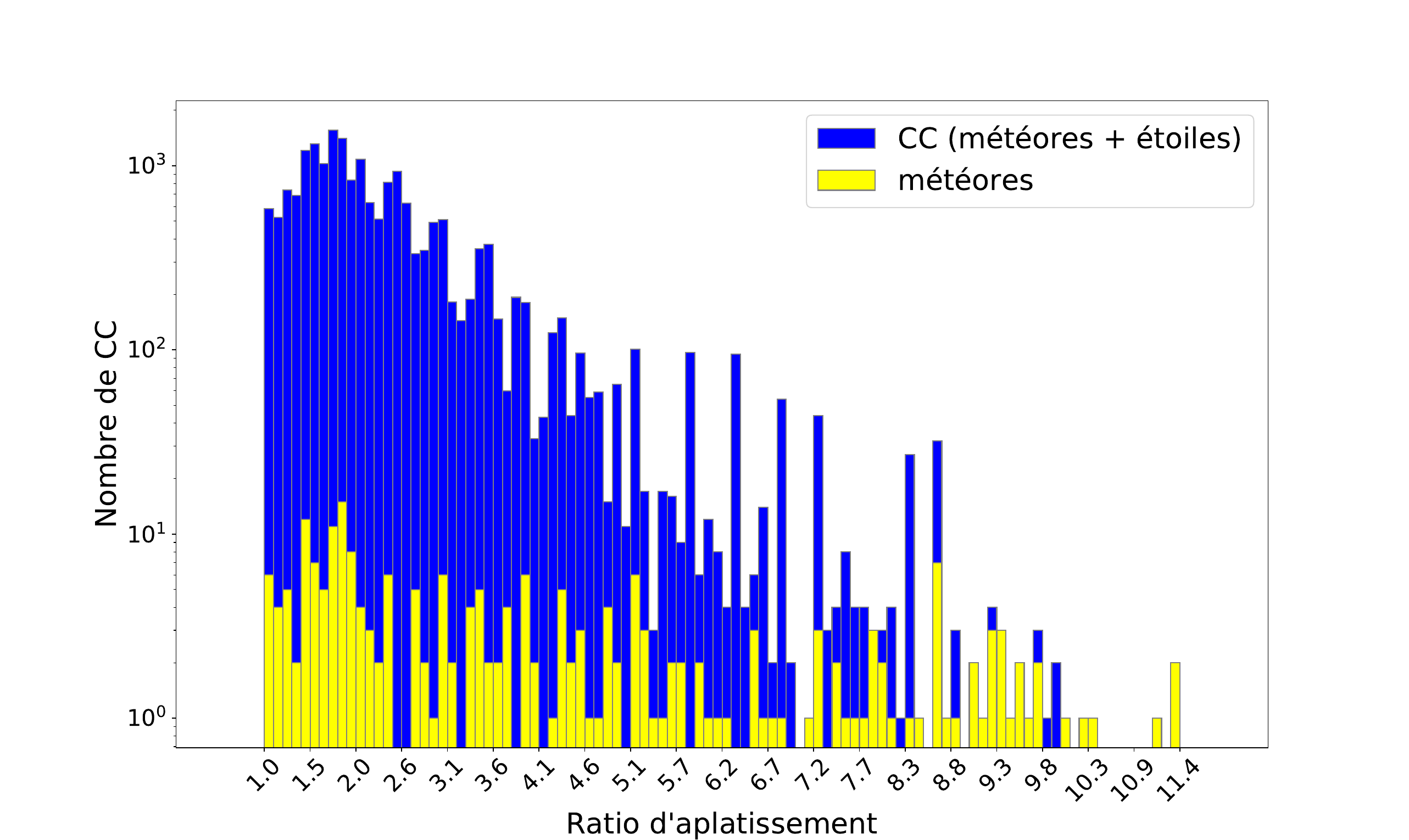}
        \vspace{-0.2cm}
        \subcaption{Exemple de séquence des $\tau$-Herculides.}
    \end{subfigure}

\end{subfigure}\hfill
\caption{Histogrammes d'aplatissement des ellipses.}
\label{fig:histo}
\vspace{-0.4cm}
\end{figure}

 

    Il apparaît \textit{a posteriori} que la détection d'ellipse fait chuter le taux de détection. Il y a deux raisons à cela. Premièrement, car la vitesse apparente générée par la caméra peut être plus importante que celle des météores et donc que l'étalement des ellipses est majoritairement dû au mouvement de la caméra. \new{Utiliser un modèle d'ellipse pour classifier les régions en fonction de leur aplatissement ne représente pas une solution dans ce cas  (Fig.~\ref{fig:histo})}. Deuxièmement, dans le cas de régions de faible luminosité, mais supérieure au seuil bas de l'hystérésis, celles-ci seront conservées, mais leur binarisation donnera lieu à des traces très fines et donc à un ratio $\rho=a/b$ plus élevé car le petit rayon $b$ sera numériquement inférieur aux régions fortement lumineuses. Cela est vrai à la fois pour les météores et les étoiles (Fig~\ref{fig:maxred}). 


\begin{table*}[htb]
    \caption{\label{stats}Statistiques sur l'ensemble de vidéos du banc de test (FP = faux positifs, FN = faux negatifs, VT = vérité terrain).}
    \begin{center}
    \vspace{-0.5cm}
    {\small
    \begin{tabular}{l r r r r r r r r r}
        \toprule
        & \multicolumn{2}{c}{Météores détectés} & \multicolumn{2}{c}{Images détectées} & \multicolumn{2}{c}{Taux (FMDT/VT)} & & \\ \cmidrule(lr){2-3} \cmidrule(lr){4-5} \cmidrule(lr){6-7}
        Séquences vidéos & FMDT & VT & FMDT & VT  & Météores & Images & FP & FN & F-score \\ \midrule
       Géminides                   &  61 &  70 &  776 & 1222 &  87\% & 64\% &  41 & 9  & 0,71 \\
       $\tau$-Herculides (Basler)  &  11 &  14 &  126 &  150 &  79\% & 84\% &  48 & 3  & 0,30 \\ 
       $\tau$-Herculides (Sony)    & 111 & 139 &  807 & 1457 &  80\% & 55\% & 130 & 28 & 0,58 \\
       $\tau$-Herculides (cluster) &  34 &  34 &  223 &  249 & 100\% & 89\% &   4 & 0  & 0,94 \\ \midrule
        Total \& moyenne           & 217 & 257 & 1932 & 3078 &  84\% & 63\% & 223 & 40 &  0,62\\ \bottomrule
    \end{tabular}
    }
    \end{center}
\end{table*}
   

\vspace{-0.1cm}
\subsection{Perspectives d'évolution}
\vspace{-0.1cm}

D'un point de vue qualité, la recherche d'ellipse très aplaties  ne permet pas de dissocier les étoiles des météores.  La solution envisagée, qui entraîne un surcoût, est de réaliser une compensation du mouvement global de l'image (par flot optique mono-échelle) avant de calculer la max-réduction.

\new{Afin de simplifier l'utilisation de FMDT, une exploration automatique de l'espace des paramètres -- s'appuyant sur une méthode d'apprentissage statistique -- est envisagée.}

Enfin, d'un point de vue temps de traitement, l'hystérésis (ECC + ACC) représente 95 \% du temps total de calcul. Un véritable algorithme d'hystérésis en une passe est en cours de développement. Il utilisera des routines en SIMD \cite{Lemaitre2020_SIMD_RLE_FLSL_WPMVP} qui n'impactent pas trop la consommation et éventuellement le multi-threading \cite{Cabaret2018_parallel_LSL_JRTIP} qui a un impact plus important.

\vspace{-0.2cm}
\section{Conclusion}
\vspace{-0.1cm}

Cet article introduit un outil de détection automatique de météores (FMDT) pour systèmes aéroportés. 

    
    \new{Avec un taux de détection de 84\%, FMDT est comparable aux résultats presentés dans l'{\'E}tat de l'Art, alors que le nombre de faux positif est considérablement inférieur à certaines chaînes. De plus, FMDT respecte les contraintes de temps réel (25 FPS) et de basse consommation (6W) sur un \rpi, contrairement aux applications utilisant les réseaux des neurones (0.2 FPS et 140W). }

    La suite du projet est d'accélérer l'hystérésis, afin de réduire le temps de traitement total et ainsi pouvoir ajouter de nouveaux algorithmes et améliorer la qualité de détection, tout en respectant les contraintes.


\vspace{-0.1cm}
\begin{small}
\bibliographystyle{plainnat-fr}
\bibliography{biblio}

\begin{thebibliography}{10}
\expandafter\ifx\csname fonteauteurs\endcsname\relax
\def\fonteauteurs{\scshape}\fi

\bibitem{Arai2014}
T.~\bgroup\fonteauteurs\bgroup Arai\egroup\egroup{},
  M.~\bgroup\fonteauteurs\bgroup Kobayashi\egroup\egroup{},
  M.~\bgroup\fonteauteurs\bgroup Yamada\egroup\egroup{},
  T.~\bgroup\fonteauteurs\bgroup Matsui\egroup\egroup{} et Cometss~Project
  \bgroup\fonteauteurs\bgroup Team\egroup\egroup{} :
\newblock Meteor observation {HDTV} camera onboard the {I}nternational {S}pace
  {S}tation.
\newblock \emph{In} {\em {LPSC}}, 2014.

\bibitem{2021ascl.soft04011A}
Y.~\bgroup\fonteauteurs\bgroup {Audureau}\egroup\egroup{} :
\newblock {Freeture: Free software to capTure meteors}.
\newblock Astrophysics Source Code Library, record ascl:2104.011, 2021.

\bibitem{Cabaret2014_CCL_SIPS}
L.~\bgroup\fonteauteurs\bgroup Cabaret\egroup\egroup{} et
  L.~\bgroup\fonteauteurs\bgroup Lacassagne\egroup\egroup{} :
\newblock What is the world's fastest connected component labeling algorithm ?
\newblock \emph{In} {\em {IEEE} International Workshop on Signal Processing
  Systems ({SiPS})}, pages 97--102, 2014.

\bibitem{Cabaret2018_parallel_LSL_JRTIP}
L.~\bgroup\fonteauteurs\bgroup Cabaret\egroup\egroup{},
  L.~\bgroup\fonteauteurs\bgroup Lacassagne\egroup\egroup{} et
  D.~\bgroup\fonteauteurs\bgroup Etiemble\egroup\egroup{} :
\newblock {P}arallel {L}ight {S}peed {L}abeling for connected component
  analysis on multi-core processors.
\newblock {\em Journal of Real Time Image Processing},
  15(1)\string:\penalty500\relax 173--196, 2018.

\bibitem{cecil}
D.~\bgroup\fonteauteurs\bgroup {Cecil}\egroup\egroup{} et
  M.~\bgroup\fonteauteurs\bgroup {Campbell-Brown}\egroup\egroup{} :
\newblock {The application of convolutional neural networks to the automation
  of a meteor detection pipeline}.
\newblock {\em Planetary and Space Science}, 186\string:\penalty500\relax
  104920, 2020.

\bibitem{Colas2020_FRIPON_short}
F.~\bgroup\fonteauteurs\bgroup Colas\egroup\egroup{} \emph{et~al.} :
\newblock Fripon: a worldwide network to track incoming meteoroids.
\newblock {\em Astronomy and Astrophysics}, 644\string:\penalty500\relax 1--23,
  2020.

\bibitem{duda1972use}
R.~O. \bgroup\fonteauteurs\bgroup Duda\egroup\egroup{} et P.~E.
  \bgroup\fonteauteurs\bgroup Hart\egroup\egroup{} :
\newblock Use of the hough transformation to detect lines and curves in
  pictures.
\newblock {\em CACM}.

\bibitem{galindo}
Y.~\bgroup\fonteauteurs\bgroup Galindo\egroup\egroup{} et
  A.~\bgroup\fonteauteurs\bgroup Lorena\egroup\egroup{} :
\newblock Deep transfer learning for meteor detection.
\newblock \emph{In} {\em ENIAC}, 2018.

\bibitem{grl}
P.~\bgroup\fonteauteurs\bgroup Gural\egroup\egroup{} et
  D.~\bgroup\fonteauteurs\bgroup Segon\egroup\egroup{} :
\newblock A new meteor detection processing approach for observations collected
  by the croatian meteor network ({CMN}).
\newblock {\em WGN, Journal of the IMO}, 37, 2009.

\bibitem{dlgural}
P.~S. \bgroup\fonteauteurs\bgroup {Gural}\egroup\egroup{} :
\newblock {Deep learning algorithms applied to the classification of video
  meteor detections}.
\newblock {\em MNRAS}, 489(4)\string:\penalty500\relax 5109--5118, 2019.

\bibitem{Lacassagne2009_LSL_ICIP}
L.~\bgroup\fonteauteurs\bgroup Lacassagne\egroup\egroup{} et
  B.~\bgroup\fonteauteurs\bgroup Zavidovique\egroup\egroup{} :
\newblock {L}ight {S}peed {L}abeling for {RISC} architectures.
\newblock \emph{In} {\em International Conference on Image Analysis and
  Processing}, pages 3245--3248. IEEE, 2009.

\bibitem{Lemaitre2020_SIMD_RLE_FLSL_WPMVP}
F.~\bgroup\fonteauteurs\bgroup Lemaitre\egroup\egroup{},
  A.~\bgroup\fonteauteurs\bgroup Hennequin\egroup\egroup{} et
  L.~\bgroup\fonteauteurs\bgroup Lacassagne\egroup\egroup{} :
\newblock How to speed {C}onnected {C}omponent {L}abeling up with {SIMD} {RLE}
  algorithms.
\newblock \emph{In} {\em Workshop on Programming Models for {SIMD}/Vector
  Processing}. ACM, 2020.

\bibitem{liegibel2022meteor}
M.~\bgroup\fonteauteurs\bgroup Liegibel\egroup\egroup{},
  J.~\bgroup\fonteauteurs\bgroup Petri\egroup\egroup{},
  P.~\bgroup\fonteauteurs\bgroup Hoffmann\egroup\egroup{},
  N.~\bgroup\fonteauteurs\bgroup Geier\egroup\egroup{} et
  S.~\bgroup\fonteauteurs\bgroup Klinkner\egroup\egroup{} :
\newblock Meteor observation with the source cubesat--developing a simulation
  to test on-board meteor detection algorithms.
\newblock \emph{In} {\em SSEA}. ESA, 2022.

\bibitem{Millet2022_Meteorix_COSPAR}
M.~\bgroup\fonteauteurs\bgroup Millet\egroup\egroup{},
  N.~\bgroup\fonteauteurs\bgroup Rambaux\egroup\egroup{},
  A.~\bgroup\fonteauteurs\bgroup Cassagne\egroup\egroup{},
  M.~\bgroup\fonteauteurs\bgroup Bouyer\egroup\egroup{},
  A.~\bgroup\fonteauteurs\bgroup Petreto\egroup\egroup{} et
  L.~\bgroup\fonteauteurs\bgroup Lacassagne\egroup\egroup{} :
\newblock High performance computer vision application for {M}eteor detection
  from a cubesat.
\newblock \emph{In} {\em Assembly of Committee on Space Research ({COSPAR})},
  2022.

\bibitem{metrec}
S.~\bgroup\fonteauteurs\bgroup Molau\egroup\egroup{} :
\newblock The meteor detection software metrec.
\newblock -1\string:\penalty500\relax 131, 01 1999.

\bibitem{MolGur}
S.~\bgroup\fonteauteurs\bgroup {Molau}\egroup\egroup{} et P.~S.
  \bgroup\fonteauteurs\bgroup {Gural}\egroup\egroup{} :
\newblock {A review of video meteor detection and analysis software}.
\newblock {\em WGN, Journal of the IMO}, 33(1)\string:\penalty500\relax 15--20,
  2005.

\bibitem{Mucherino2009}
A.~\bgroup\fonteauteurs\bgroup Mucherino\egroup\egroup{}, P.~J.
  \bgroup\fonteauteurs\bgroup Papajorgji\egroup\egroup{} et P.~M.
  \bgroup\fonteauteurs\bgroup Pardalos\egroup\egroup{} :
\newblock {\em k-Nearest Neighbor Classification}, pages 83--106.
\newblock Springer, 2009.

\bibitem{ocana2019balloon}
F.~\bgroup\fonteauteurs\bgroup Oca{\~n}a\egroup\egroup{},
  A.~\bgroup\fonteauteurs\bgroup {S\'anchez de Miguel}\egroup\egroup{},
  Daedalus \bgroup\fonteauteurs\bgroup Project\egroup\egroup{} \emph{et~al.} :
\newblock Balloon-borne video observations of geminids 2016.
\newblock {\em preprint arXiv:1911.10064}, 2019.

\bibitem{Petreto2018_SIMD_GPU_OF_DASIP}
A.~\bgroup\fonteauteurs\bgroup Petreto\egroup\egroup{},
  A.~\bgroup\fonteauteurs\bgroup Hennequin\egroup\egroup{},
  \bgroup\fonteauteurs\bgroup \egroup\egroup{}, T.~\bgroup\fonteauteurs\bgroup
  Koehler\egroup\egroup{}, T.~\bgroup\fonteauteurs\bgroup
  Romera\egroup\egroup{}, Y.~\bgroup\fonteauteurs\bgroup
  Fargeaix\egroup\egroup{}, B.~\bgroup\fonteauteurs\bgroup
  Gaillard\egroup\egroup{}, M.~\bgroup\fonteauteurs\bgroup
  Bouyer\egroup\egroup{}, Q.~L. \bgroup\fonteauteurs\bgroup
  Meunier\egroup\egroup{} et L.~\bgroup\fonteauteurs\bgroup
  Lacassagne\egroup\egroup{} :
\newblock Energy and execution time comparison of optical flow algorithms on
  {SIMD} and {GPU} architectures.
\newblock \emph{In} {\em DASIP}. IEEE, 2018.

\bibitem{Rambaux2019_ESA_short}
N.~\bgroup\fonteauteurs\bgroup Rambaux\egroup\egroup{} \emph{et~al.} :
\newblock Meteorix: a cubesat mission dedicated to the detection of meteors and
  space debris.
\newblock \emph{In} {\em ESA Space Safety Programme Office, NEO and Debris
  Detection Conference}.

\bibitem{Vaubaillon2023_short}
J.~\bgroup\fonteauteurs\bgroup Vaubaillon\egroup\egroup{} \emph{et~al.} :
\newblock A 2022 $\tau$-herculid meteor cluster from an airborne experiment:
  Automated detection, characterization, and consequences for meteoroids.
\newblock {\em Astronomy and Astrophysics}, 2023.

\end{thebibliography}
\end{small}
\end{sloppypar}
\end{document}